\documentclass[10pt,twocolumn,letterpaper]{article}

\usepackage{cvpr}
\usepackage{times}
\usepackage{epsfig}
\usepackage{graphicx}
\usepackage{amsmath}
\usepackage{amssymb}
\usepackage{color}
\usepackage{float}

\usepackage{algorithm}
\usepackage[noend]{algpseudocode}
\usepackage[english]{babel}
\usepackage[T1]{fontenc}
\usepackage[utf8]{inputenc}
\usepackage{booktabs} 
\usepackage{colortbl}
\usepackage{comment}

\usepackage[english]{babel}
\usepackage[T1]{fontenc}
\usepackage[utf8]{inputenc}
\usepackage{enumitem}
\setlist{nosep}

\usepackage{times,color}
\usepackage{parskip}
\usepackage{ifthen}
\usepackage{float}
\usepackage{alltt}
\usepackage{mathenv}
\usepackage{amsmath}
\usepackage{amssymb}
\usepackage{amsthm}
\usepackage{newlfont} 
\usepackage{floatflt}
\usepackage{wrapfig}
\usepackage{fixltx2e}
\usepackage{subfig} 
\usepackage{multirow}

\newcommand{\code}{\mathbf{z}}

\newcommand{\CodecEncoder}{\mathrm{E}}
\newcommand{\CodecDecoder}{\mathrm{D}}

\newcommand{\RegressorName}{{I2ZNet}}
\newcommand{\Image}{\mathbf{I}}
\newcommand{\HeadPose}{\mathbf{H}}
\newcommand{\CFTC}{\textrm{CFTC}}
\newcommand{\MOTC}{\textrm{MOTC}}
\newcommand{\FLRC}{\textrm{FLRC}}

\newcommand{\DA}{\textrm{DA}}

\definecolor{RedColor}{rgb}{0.7,0,0} 
\definecolor{GreenColor}{rgb}{0,0.4,0} 
\definecolor{BlueColor}{rgb}{0,0,0.8} 
\definecolor{CyanColor}{rgb}{0,1,1} 
\newcommand{\takaaki}[1]{{\color{RedColor} Takaaki: #1 $\qed$}}
\newcommand{\shooui}[1]{{\color{GreenColor} Shoou-I: #1 $\qed$}}

\newcommand{\hyunsoo}[1]{{\color{CyanColor} Hyun Soo: #1 $\qed$}}

\usepackage[pagebackref=true,breaklinks=true,letterpaper=true,colorlinks,bookmarks=false]{hyperref}

\cvprfinalcopy 

\ifcvprfinal\fi

\cvprfinalcopy 

\cvprfinalcopy

\usepackage[bbgreekl]{mathbbol}
\usepackage{amsfonts}
\DeclareSymbolFontAlphabet{\mathbb}{AMSb}
\DeclareSymbolFontAlphabet{\mathbbl}{bbold}

\begin{document}


\title{Self-Supervised Adaptation of High-Fidelity Face Models for \\ Monocular Performance Tracking}

\author{
Jae Shin Yoon$^\dagger$
\hspace{10mm}Takaaki Shiratori$^\ddagger$
\hspace{10mm}Shoou-I Yu$^\ddagger$
\hspace{10mm}Hyun Soo Park$^\dagger$
\\
\hspace{0mm}$^\dagger$University of Minnesota
\hspace{20mm}
$^\ddagger$Facebook Reality Labs
\\
{\tt\small \hspace{8mm}\{jsyoon, hspark\}@umn.edu \hspace{6mm} \{tshiratori, shoou-i.yu\}@fb.com}
}



\twocolumn[{%
\maketitle
\thispagestyle{empty}
	\begin{center}
		\centering
	\includegraphics[width=7in]{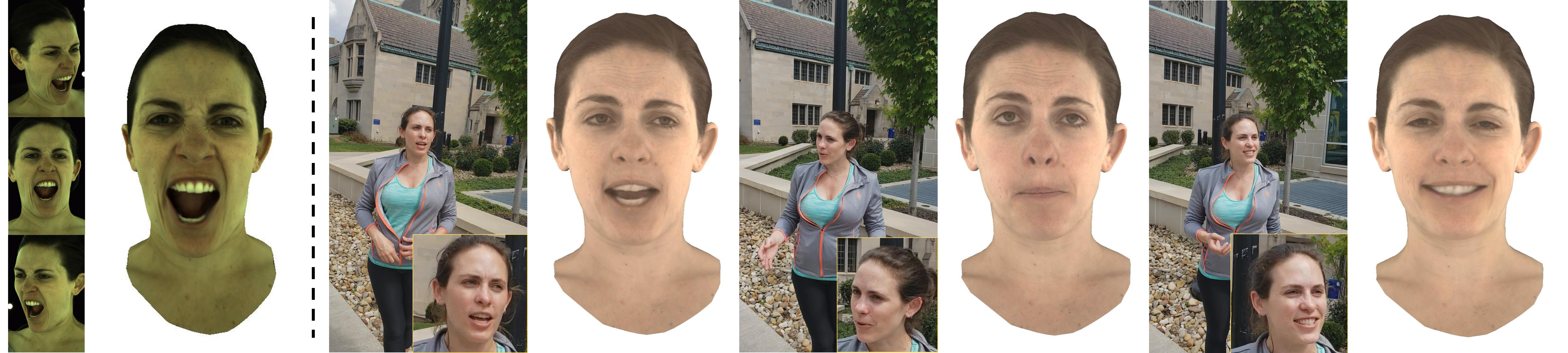}
	\vspace{-6mm}
	\captionof{figure}{Results of high-fidelity 3D facial performance tracking from our method, which automatically adapts a high-quality face model~\cite{Lombardi_SIGGRAPH2018} captured in a controlled lab environment (left) to in-the-wild imagery (right) through our proposed self-supervised domain adaptation method. Note the fine details we are able to recover from cellphone quality video.}
	\label{fig:teaser}
	\vspace{5mm}
\end{center}	
\label{fig:teaser_main}
}]

\begin{abstract}
\vspace{-2mm}

Improvements in data-capture and face modeling techniques have enabled us to create high-fidelity realistic face models. However, driving these realistic face models requires special input data, \eg 3D meshes and unwrapped textures. Also, these face models expect clean input data taken under controlled lab environments, which is very different from data collected in the wild. All these constraints make it challenging to use the high-fidelity models in tracking for commodity cameras. In this paper, we propose a self-supervised domain adaptation approach to enable
the animation of high-fidelity face models from a commodity camera.
Our approach first circumvents the requirement for special input data by training a new network
that can directly drive a face model just from a single 2D image. Then, we overcome the domain mismatch between lab and uncontrolled environments
by performing self-supervised domain adaptation based on ``consecutive frame texture consistency'' based on the assumption that the appearance of the face is consistent over consecutive frames,
avoiding the necessity of modeling the new environment such as lighting or background.
Experiments show that we are able to drive a high-fidelity face model to perform complex facial motion from a cellphone camera without requiring
any labeled data from the new domain.

\vspace{-4mm}

\end{abstract}

\section{Introduction}
\label{sec:Introduction}

High-fidelity face models enable the building of realistic avatars, which play a key role in communicating ideas, thoughts and emotions.
Thanks to the uprising of data-driven approaches,
highly realistic and detailed face models can be created
with active appearance models (AAMs)~\cite{cootes1995active, cootes2001active}, 
3D morphable models (3DMMs)~\cite{blanz1999morphable},
or deep appearance models (DAMs)~\cite{Lombardi_SIGGRAPH2018}.
These data-driven approaches
jointly model facial geometry and appearance, thus empowering the model
to learn the correlation between the two and synthesize high-quality facial images.
Particularly, the recently proposed DAMs can model and generate realistic animation and view-dependent textures with pore-level details by leveraging the high capacity of deep neural networks.

Unfortunately, barriers exist when applying these high-quality models to monocular in-the-wild imagery
due to \textit{modality mismatch} and \textit{domain mismatch}.
Modality mismatch refers to the fact that 
high-fidelity face modeling and tracking requires specialized input data, 
(\eg DAMs require tracked 3D meshes and unwrapped textures) which is not easily accessible on consumer-grade mobile capture devices.
Domain mismatch refers to the fact that 
the visual statistics of in-the-wild imagery are considerably different from that of a 
controlled lab environment used to build the high-fidelity face model.
In-the-wild imagery includes various background clutter, low resolution, and complex ambient lighting. 
Such domain gap breaks the correlation between appearance and geometry learned by the data-driven model
and the model may no longer work well in the new domain.
The existence of these two challenges greatly inhibits the wide-spread use of
the high-fidelity face models.


In this paper, we present a method to perform high-fidelity face tracking for monocular in-the-wild imagery based on DAMs face model learned
from lab-controlled imagery.
Our method bridges the controlled lab domain and in-the-wild domain 
such that we can perform high-fidelity face tracking with DAM face models on in-the-wild video camera sequences. 
To tackle the modality mismatch, we train \RegressorName, a deep neural network that takes a monocular image as input
and directly regresses to the intermediate representation of the DAM, thus circumventing the need
for 3D meshes and unwrapped textures required in DAMs.
As \RegressorName$\,$ relies on data captured in a lab environment and cannot handle the domain mismatch,
we present a self-supervised domain adaptation technique that can adapt \RegressorName$\,$
to new environments without requiring any labeled data from the new domain.
Our approach leverages the assumption that the textures (appearance) of a face between consecutive frames should be consistent 
and incorporates this source of supervision to adapt the domain of \RegressorName$\,$ such that 
final tracking results preserve consistent texture over consecutive frames on target-domain imagery, as shown in Figure~\ref{fig:teaser}.
The resulting face tracker outperforms state-of-the-art face tracking methods in terms of geometric accuracy, temporal stability, and visual plausibility. 

The key strength of this approach is that we do not make any other assumptions on the scene or
lighting of in-the-wild imagery, enabling our method to be applicable to a wide 
variety of scenes.
Furthermore, our method computes consistency for all visible portions of the texture, thus
providing significantly more supervision and useful gradients than per-vertex based methods\,\cite{tewari2017self,dong2018supervision}.
Finally, we emphasize that the consecutive frame texture consistency assumption is not simply a regularizer to avoid overfitting.
This assumption provides an additional source of supervision which enables our model
to adapt to new environments and achieve considerable improvement of accuracy and stability.
\vspace{-1mm}

In summary, the contributions of this paper are as follows:
\vspace{-1mm}
\begin{enumerate}
    \item \RegressorName, a deep neural network that can directly predict the intermediate representation of a DAM from a single image.
    \item A self-supervised domain adaptation method based on consecutive frame texture consistency to enhance face tracking. No labeled data is required for images from the target domain.
    \item High-fidelity 3D face tracking on in-the-wild videos captured with a commodity camera.
\end{enumerate}
\section{Related Work}
\label{sec:RelatedWork}
Humans are evolved to decode, understand and convey non-verbal information from facial motion, \eg, a subtle unnatural eye blink, symmetry, and reciprocal response can be easily detected. Therefore, the realistic rendering of facial motion is key to enable telepresence technology~\cite{Lombardi_SIGGRAPH2018}. This paper lies in the intersection between high fidelity face modeling and 3D face reconstruction from a monocular camera, which will be briefly reviewed here.

\noindent\textbf{3D Face Modeling} Faces have underlying spatial structural patterns where low dimensional embedding can efficiently and compactly represent diverse facial configurations, shapes, and textures. Active Shape Models (ASMs)~\cite{cootes1995active} have shown strong expressibility and flexibility to describe a variety of facial configurations by leveraging a set of facial landmarks. However, the nature of the sparse landmark dependency limits the reconstruction accuracy that is fundamentally bounded by the landmark localization. 
AAMs~\cite{cootes2001active} address the limitation by exploiting the photometric measure using both shape and texture, resulting in compelling face tracking. Individual faces are combined into a single 
3DMM~\cite{blanz1999morphable} by computing dense correspondences based on optical flow in conjunction with the shape and texture priors in a linear subspace. Large-scale face scans (more than 10,000 people) from diverse population enables modeling of accurate distributions of faces~\cite{booth2018large,booth20173d}. With the aid of multi-camera systems and deep neural networks, the limitation of the linear models can be overcome using DAMs~\cite{Lombardi_SIGGRAPH2018} that predicts high quality geometry and texture. Its latent representation is learned by a conditional variational autoencoder~\cite{kingma2013auto} that encodes view-dependent appearance from different viewpoints. 
Our approach eliminates the multi-camera requirement of the DAMs by adapting the networks to a video from a monocular camera.

\noindent\textbf{Single View Face Reconstruction} The main benefit of the compact representation of 3D face modeling is that it allows estimating the face shape, appearance, and illumination parameters from a single view image. 
For instance, the latent representation of the 3DMMs can be recovered by jointly optimizing pixel intensity, edges and illumination (approximated by spherical harmonics)~\cite{romdhani2005estimating}. The recovered 3DMMs can be further refined to fit to a target face using a collection of photos~\cite{roth2016adaptive} or depth based camera~\cite{cao2014facewarehouse}.
\cite{tewari2017mofa} leveraged expert designed rendering layers which model face shape, expression, and illumination and utilized inverse rendering to estimate a set of compact parameters which renders a face that best fits the input. 
This is often an simplification and cannot model all situations. 
In contrast, our method does not make any explicit assumptions on the lighting of the scene, and thus achieves more flexibility to different environments.

Other methods include 
\cite{jeni2015dense, zhu2017face}, which used cascaded CNNs which densely align the 3DMM with a 2D face in an iterative way based on facial landmarks.
The geometry of a 3D face is regressed in a coarse-to-fine manner~\cite{richardson2017learning}, and asymmetric loss enforces the network to regress the identity consistent 3D face~\cite{tran2017regressing}. 
\cite{tewari2017self} utilizes jointly learned geometry and reflectance correctives to fit in-the-wild faces. \cite{feng2018joint} trained UV regression maps which jointly align with the 3DMM to directly reconstruct a 3D face.

\noindent\textbf{Tackling Domain Mismatch} A key challenge is the oftentimes significant gap between the distribution of training and testing data.
To this end, \cite{richardson20163d, tran2017regressing} utilized synthetic data to boost 3D face reconstruction performance.
A challenge here is to generate synthetic data that is representative of the testing distribution.
\cite{dong2018supervision} utilized domain invariant motion cues to perform unsupervised domain adaptation
for facial landmark tracking. 
While their method was tested on \textit{sparse} landmarks and benefited from a limited source of supervision,
our method performs \textit{dense} per-pixel matching of textures,  providing more supervision for domain adaptation.

\begin{figure}[t]
	\begin{center}
		\includegraphics[width=3.4in]{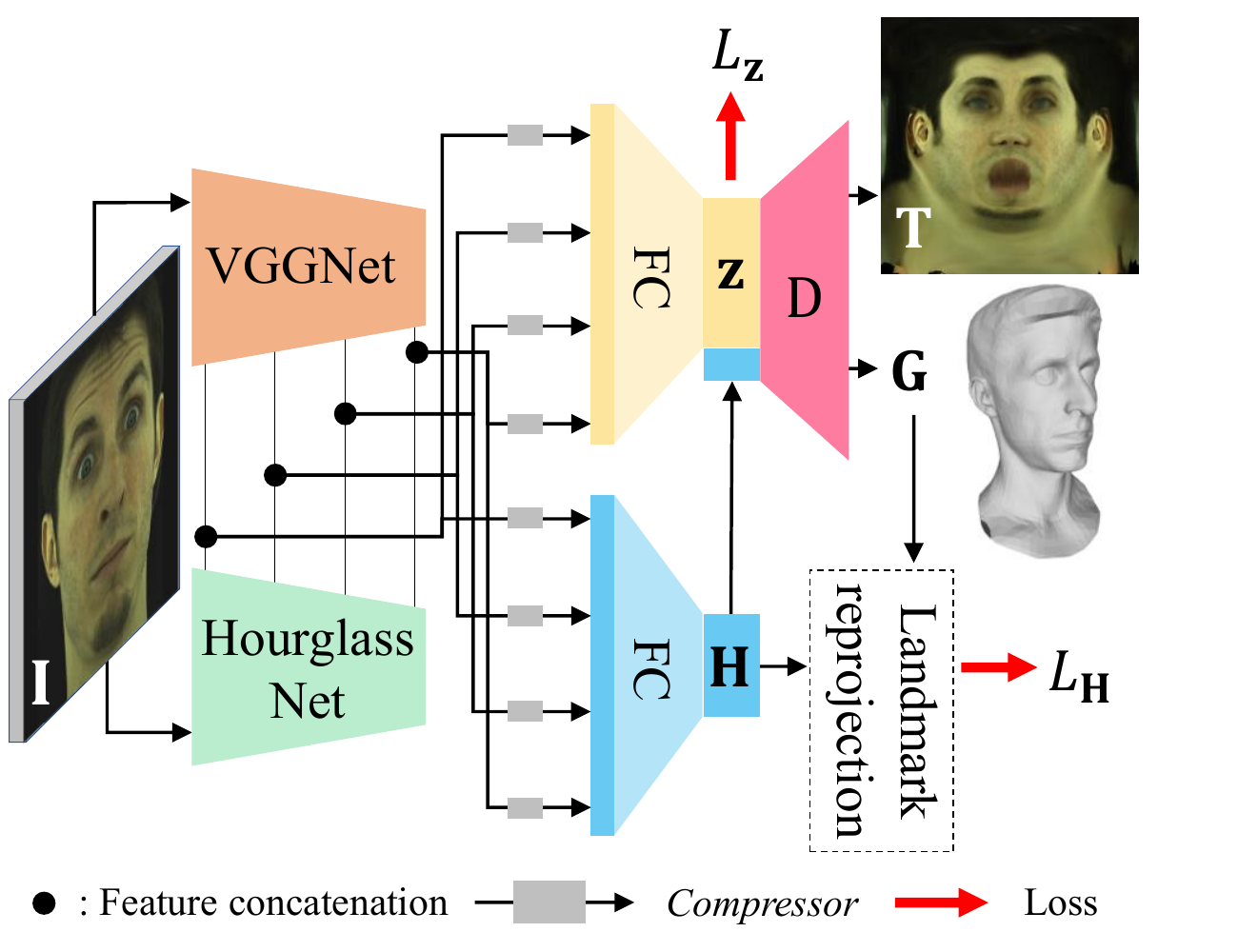}
	\end{center}	
	\vspace{-2mm}
	\caption{Illustration of the \RegressorName\ architecture. 
	\RegressorName\ extracts the domain-invariant perceptual features and facial image features using the pre-trained VGGNet~\cite{simonyan2014very} and HourglassNet~\cite{newell2016stacked}, respectively, from an input image $\Image$. The combined multiple depth-level features are then regressed to the latent code \textbf{z} of the pre-trained DAMs ($\CodecDecoder$)~\cite{Lombardi_SIGGRAPH2018} and the head pose \textbf{H} via fully connected layers. 
	I2ZNet is trained with the losses defined for $\code$ and $\HeadPose$, namely $L_\code$ and $L_\HeadPose$, as well as the view consistency loss in Eq.~\eqref{eq:I2ZLoss_ViewConsistency}.
	}
	\label{fig:network_main}
\end{figure}

\section{Methodology}

When applying existing face models such as AAMs and DAMs to monocular video recordings,
we face two challenges: modality mismatch and domain mismatch.
Modality mismatch occurs when the existing face model requires input data to be represented in a face centric representation 
such as 3D meshes with unwrapped texture in a pre-defined topology. 
This representation does not comply with an image centric representation, thus preventing us from using these face models.
Domain mismatch occurs when the visual statistics of in-the-wild images are different from that of the scenes used to construct the models. In the following sections, we first present \RegressorName$\,$ for the modality mismatch, 
and then describe how to adapt \RegressorName$\,$ in a self-supervised fashion for the domain mismatch.



\subsection{Handling Modality Mismatch}
\label{sec:handling_modality_mismatch}

Many face models including DAMs can be viewed as an encoder and decoder framework.
The encoder $\CodecEncoder_X$ takes an input $X = (\mathbf{G}, \mathbf{T})$,
which corresponds to the geometry and unwrapped texture, respectively.
$\mathbf{G} \in \mathbb{R}^{G \times 3}$ represents the 3D locations of $G$ vertices
which form a 3D mesh of the face. 
Note that rigid head motion has already been removed from the vertex locations,
\ie $\mathbf{G}$ represents only local deformations of the face.
The unwrapped texture $\mathbf{T} \in \mathbb{R}^{T \times T \times 3}$ 
is a 2D image 
that represents the appearance at different locations on $\mathbf{G}$ in the UV space.
The output of $\CodecEncoder_X$ is the intermediate code $\code$.
The decoder $\CodecDecoder$ then takes $\code$ and computes a reconstructed output 
$\widetilde{X} = \CodecDecoder(\code) = \CodecDecoder(\CodecEncoder_X(X))$.
The encoder and decoder are learned by minimizing the difference between $X$ and $\widetilde{X}$
for a large number of training samples.

The challenge is that $X = (\mathbf{G}, \mathbf{T})$, \ie, the 3D geometry and unwrapped texture, is not readily available in a monocular image $\Image$.
Therefore, we learn a separate deep encoder called \textit{\RegressorName} $\,$(Image-to-$\code$ network):
$(\code, \HeadPose) \leftarrow \CodecEncoder_\Image(\Image)$, which takes a monocular image \textbf{I} as input
and directly outputs $\code$ and the rigid head pose $\HeadPose$. 
\RegressorName\ first extracts the domain independent two-stream features using the pre-trained VGGNet~\cite{simonyan2014very} and HourglassNet~\cite{newell2016stacked}, which provides perceptual information and facial landmarks, respectively. 
The multiple depth-level two-stream features are combined with skip connections, and are regressed respectively to the intermediate representation $\mathbf{z}\in \mathbb{R}^{128}$ and the head pose $\mathbf{H}\in \mathbb{R}^{6}$ using several  fully connected layers~\cite{yoon2017pixel}.
This architecture allows to directly predicts the parameters (\textbf{z}, \textbf{H}) based on the category-level semantic information from the deep layers and local geometric/appearance details from the shallow layers at the same time. 
$\code$ can be given to the existing decoder $\CodecDecoder$ to decode the 3D mesh and texture, while $\HeadPose$ allows to reproject the decoded 3D mesh onto the 2D image. 
Figure~\ref{fig:network_main} illustrates the overall architecture of I2ZNet, and 	more details are described in the supplementary manuscript.


$\CodecEncoder_\Image$ is trained in a supervised way with multiview image sequences used for training $\CodecEncoder_X$ and $\CodecDecoder$ of DAMs.
The by-product of learning $\CodecEncoder_X$ and $\CodecDecoder$ are the latent code $\code_{gt}$
and the head pose $\HeadPose_{gt}$ at each time.
As a result of DAM training, we acquire as many tuples of $\{ \Image_v, \code_{gt}, \HeadPose_{gt} \}$ as the camera views $\{v\}$ at every time $t$ as training data for $\CodecEncoder_\Image$.

The total loss to train $\CodecEncoder_\Image$ is defined as
\begin{align}
    L_{\text{E}_\textbf{{I}}} = \lambda_{\text{z}}L_{\text{z}}+\lambda_{\text{H}}L_{\text{H}}+\lambda_\mathrm{view}L_\mathrm{view},
\end{align}
where $L_{\text{z}}$ and $L_{\text{H}}$ are the losses for $\mathbf{z}$ and $\mathbf{H}$, respectively, and $L_\mathrm{view}$ is the view-consistency loss.
$\lambda_{\text{z}}$, $\lambda_{\text{H}}$ and $\lambda_{\mathrm{view}}$ are weights for $L_{\text{z}}$, $L_{\text{H}}$ and $L_\text{view}$, respectively. 

$L_{\text{z}}$ is the direct supervision term for $\textbf{z}$ defined as
\begin{align}
L_{\text{z}}=  \sum_{v,t} \left\| \mathbf{z}_{\Image_v^t}-\mathbf{z}_{gt}^t\right\|^{2}_{2},
\end{align}
where $\mathbf{z}_{\Image}$ is a DAM latent code regressed from $\textbf{\text{I}}$ via $\CodecEncoder_\Image$. 

Inspired by~\cite{tewari2017self, kanazawaHMR18}, 
we formulate $L_\HeadPose$ as the reprojection error of the 3D landmarks predicted via $\CodecEncoder_\Image$ w.r.t.\,the 2D ground-truth landmarks $\mathbf{K}_{gt}\in \mathbb{R}^{K\times2}$ for the head pose prediction:
\begin{align}
 L_{\text{H}}= \frac{1}{K} \sum_{k,v,t} \left\|\boldsymbol{{\Pi}}\HeadPose_{\Image_v^t}\mathbf{K}^k ( \mathbf{G}_{\Image_v^t}) -\mathbf{K}_{gt}^{k}\right\|^{2}_{2},
\end{align}
where $K$ is the number of landmarks, 
$\mathbf{\Pi}=[1~0~0 ;0~1~0]$ is a weak perspective projection matrix, 
and $\mathbf{H}_\mathbf{I}$ is the head pose regressed from $\mathbf{I}$ via \RegressorName. 
$\mathbf{G}_\Image$ is the set of vertex locations decoded from $\code_\Image$ via $\CodecDecoder$,
and $\mathbf{K}^k (\cdot)$ computes the 3D location of $k$-th landmark from $\mathbf{G}_\Image$. 

Because the training image data is captured with synchronized cameras,
we want to ensure that the regressed $\code$ is the same for images from different views captured at the same time.
Therefore, we incorporate the view-consistency loss $L_\mathrm{view}$, defined as
\begin{align}
L_\mathrm{view}=\sum_{v, w, t} \left\|\code_{\Image_v^t}-\code_{\Image_w^t}\right\|^{2}_{2}.    
\label{eq:I2ZLoss_ViewConsistency}
\end{align}
We randomly select two views at every training iteration.

\begin{figure}[t]
	\begin{center}
		\includegraphics[width=3.2in]{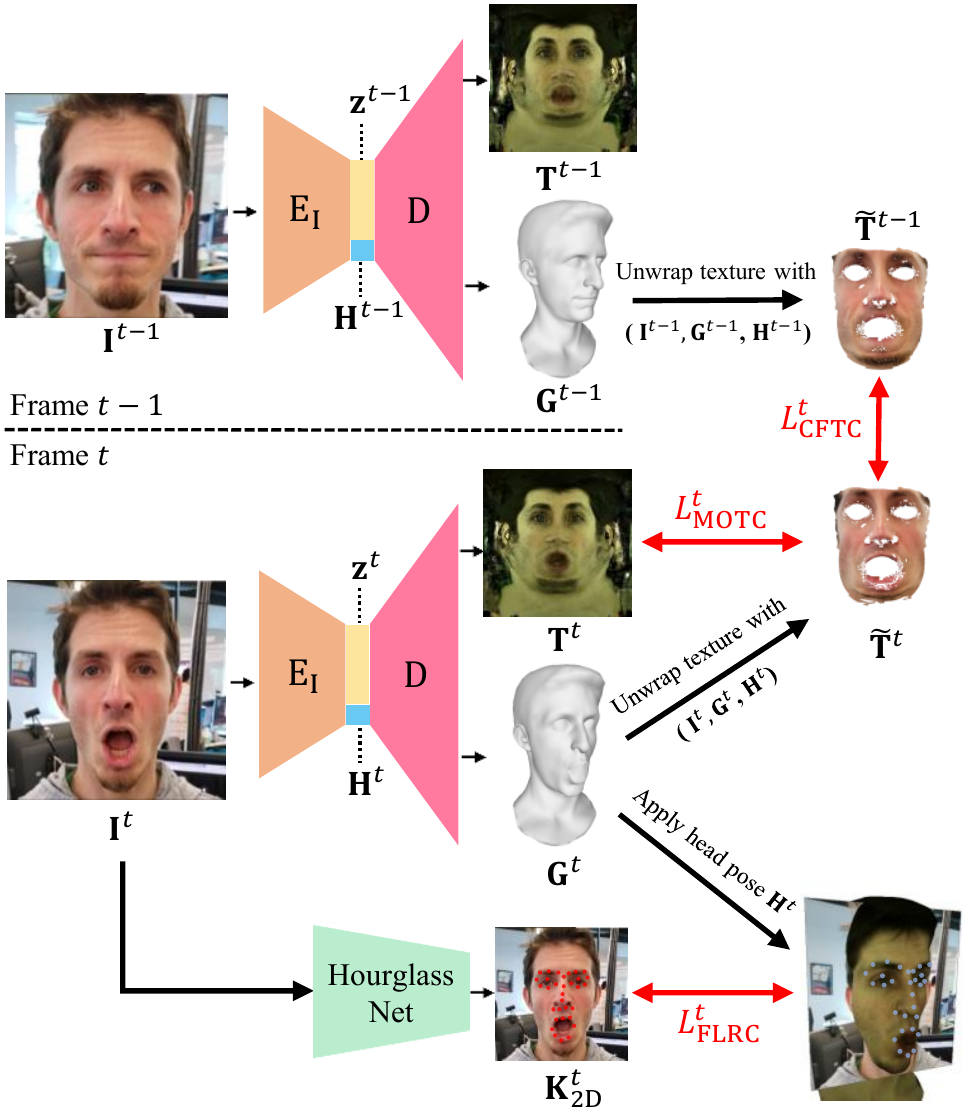}
	\end{center}	
	\vspace{-1mm}
	\caption{Overview of our self-supervised domain adaptation process. Given two consecutive frames ($\mathbf{I}^{t-1}$, $\mathbf{I}^t$), we run $\CodecEncoder_\Image$ followed by $\CodecDecoder$ to acquire the geometry ($\mathbf{G}^{t-1}$, $\mathbf{G}^t$), textures ($\mathbf{T}^{t-1}$, $\mathbf{T}^t$) and head poses ($\mathbf{H}^{t-1}$, $\mathbf{H}^t$). 
	Then, $\Image$, $\mathbf{G}$ and $\mathbf{H}$ are used to compute observed textures ($\widetilde{\mathbf{T}}^{t-1}$, $\widetilde{\mathbf{T}}^t$). These enable us to compute $L_{\textrm{CFTC}}$ and $L_{\textrm{MOTC}}$. For frame $t$, we run a hourglass facial landmark detector to get 2D landmark locations $\mathbf{K}^t_{\textrm{2D}}$, which is then used to compute $L_{\textrm{FLRC}}$. These losses can back-propagate gradients back to $\CodecEncoder_\Image$ to perform self-supervised domain adaptation.
	}
	\label{fig:fine-tuning}
\end{figure}

\subsection{Handling Domain Mismatch}
To handle the domain mismatch, we adapt \RegressorName\ 
to a new domain using a set of unlabeled images in a self-supervised manner.
The overview of the proposed domain adaptation is illustrated in Figure~\ref{fig:fine-tuning}.
Given a monocular video, we refine the encoder $\CodecEncoder_\Image$ by minimizing the domain adaptation loss $L_{\DA}$ (Eq.~\eqref{eq:TotalLoss}), which consists of (1) consecutive frame texture consistency $L_{\CFTC}$, (2) model-to-observation texture consistency $L_{\MOTC}$, and (3) facial landmark reprojection consistency $L_{\FLRC}$:
\begin{align}
L_{\DA}^t= \lambda_{\CFTC} L_{\CFTC}^t + \lambda_{\MOTC} L_{\MOTC}^{t} + \lambda_{\FLRC} L_{\FLRC}^{t},
\label{eq:TotalLoss}
\end{align}

\vspace{-3mm}
where $\lambda_{\CFTC}$, $\lambda_{\MOTC}$ and $\lambda_{\FLRC}$ correspond to the weights
for each loss term.
$L_{\CFTC}$ is our key contribution.
It adapts $\CodecEncoder_\Image$ such that textures computed from predicted geometry are temporally coherent.
$L_{\MOTC}$ enforces the consistent color of DAM generated texture with the observed texture via pixel-wise matching.
$L_{\FLRC}$ anchors the tracked 3D face by minimizing the reprojection error of the 3D model landmarks with the detected facial landmarks.




\vspace{-4mm}
\subsubsection{Consecutive Frame Texture Consistency}
\vspace{-3mm}

\begin{figure}[t]
	\begin{center}
		\includegraphics[width=3.2in]{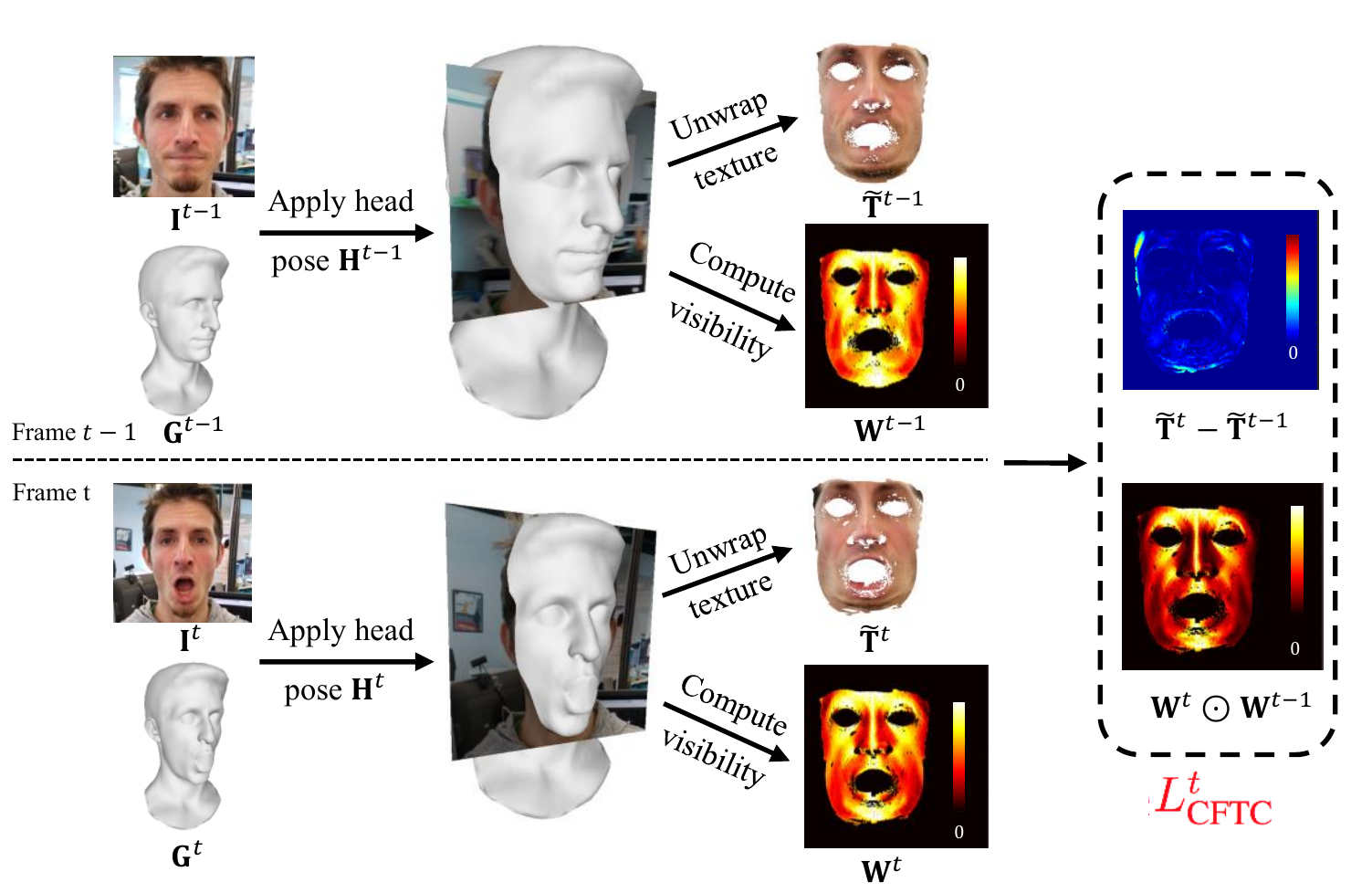}
	\end{center}	
	\vspace{-4mm}
	\caption{Illustration of how to compute $L_{\CFTC}$.
	}
	\label{fig:CFTC}
\vspace{-3mm}
\end{figure}

Inspired by the brightness constancy assumption employed in many optical flow algorithms,
we can reasonably assume that 3D face tracking for two consecutive frames is accurate only if unwrapped textures for the two frames are nearly identical.
Inversely, if we see large changes in unwrapped texture across consecutive frames,
it is highly likely due to inaccurate 3D geometry predictions. 
We make the assumption that environmental lighting and 
the appearance of the face does not change 
significantly between consecutive frames, which is satisfied in most scenarios.
Otherwise, we do not make any assumptions on the lighting
environment of a new scene, which makes our method more generalizable than existing
methods which, for example, approximates lighting with spherical harmonics \cite{tewari2017self}.

 The consecutive frame texture consistency loss $L_{\CFTC}$ is:
\begin{align}
L_{\CFTC}^{t} = \frac{1}{ W^{t,t-1} } 
\left\| 
(\mathbf{W}^{t} \odot \mathbf{W}^{t-1}) \odot 
(\widetilde{\mathbf{T}}^{t} -\widetilde{\mathbf{T}}^{t-1}) \right\|_F^{2},
\label{eq:CFTC}
\end{align}
where $\mathbf{W} \in \mathbb{R}^{T\times T}$ is a confidence matrix, $\widetilde{\mathbf{T}}$ is a texture obtained by projecting $\mathbf{G}_\Image$ onto $\Image$ with $\HeadPose_\Image$, and $\odot$ is element-wise multiplication.
We use the cosine of incident angle of the ray from the camera center to each texel  as the confidence to reduce the effect of texture distortion caused at grazing angles. 
Elements smaller than a threshold in $\mathbf{W}^{t} \odot \mathbf{W}^{t-1}$ are set to 0.
$W^{t,t-1}$ is the number of non-zero elements in $\mathbf{W}^{t} \odot \mathbf{W}^{t-1}$. 
Figure~\ref{fig:CFTC} shows example confidence matrices and textures as well as computation of $L_\CFTC$.



$\widetilde{\mathbf{T}}$ is obtained by projecting the 3D location of each texel decoded from $\code$ to an observed image $\Image$. 
\begin{align}
\label{eq:reproj_texture}
\widetilde{\mathbf{T}}_{ij} = \Image 
( \mathbf{\Pi} \HeadPose_\Image \mathbf{X}(\mathbf{G}_\Image, i,j) ),
\end{align}
where $(i,j)$ is texel coordinates.
Unlike existing methods that compute per-vertex texture loss~\cite{tewari2017self,dong2018supervision}, $L_\CFTC$ considers all visible texels,
providing significantly richer source of supervision and gradients than per-vertex-based methods.
The aforementioned steps are all differentiable, thus the entire model
can be updated in an end-to-end fashion.  

\vspace{-3mm}
\subsubsection{Model-to-Observation Texture Consistency}
\label{sec:color_correct}
\vspace{-2mm}

This loss enforces the predicted textures $\mathbf{T}$ to match the texture observed in the image $\widetilde{\mathbf{T}}$.
Although this is similar to the photometric loss used in \cite{tewari2017self},
a challenge in our technique is the aforementioned domain mismatch: 
$\mathbf{T}$ could be significantly different from $\widetilde{\mathbf{T}}$ mainly due to lighting condition changes.
Therefore, we incorporate an additional network  $\mathbf{T} \leftarrow \textrm{C}(\mathbf{T})$ to convert the color of the predicted texture to the one of the currently observed texture. 
$\textrm{C}(\mathbf{T})$ is also learned, and since training data is limited,
we learn a single 1-by-1 convolutional filter 
which can be viewed as the color correction matrix and corrects the white-balance
between the two textures. 
The model-to-observation texture consistency (MOTC) is formulated as
\begin{align}
L_{\textrm{MOTC}}^{t}= \frac{1}{W^t} 
\left\| \mathbf{W}^{t} \odot \left(\widetilde{\mathbf{T}}^{t}-\textrm{C}\left(\mathbf{T}^{t}\right) \right)\right\|_F^{2}.
\label{eq:MOTC}
\end{align}

\vspace{-5mm}
\subsubsection{Facial Landmark Reprojection Consistency}
\label{sec:tune_key}
\vspace{-3mm}
This loss enforces a sparse set of vertices on the 3D mesh corresponding to the landmark locations
to be consistent with 2D landmark predictions. 
Given $K$ facial landmarks, the facial landmark reprojection consistency (FLRC) loss is formulated as: 
\vspace{-3mm}

\begin{align}
L_{\textrm{FLRC}}^{t}= \frac{1}{K} \sum_{k}\left\|\mathbf{K}^{k,t}_{\textrm{2D}} - \mathbf{\Pi} \HeadPose_\Image^t \mathbf{K}^k({\mathbf{G}_\Image^t}) \right\|^{2}_2,
\end{align}

\vspace{-3mm}
where 
$\mathbf{K}^{k,t}_{\textrm{2D}}$ 
is the location of the $k$-th detected 2D landmark.



\begin{figure}[t]
	\begin{center}
		\includegraphics[width=3.3in]{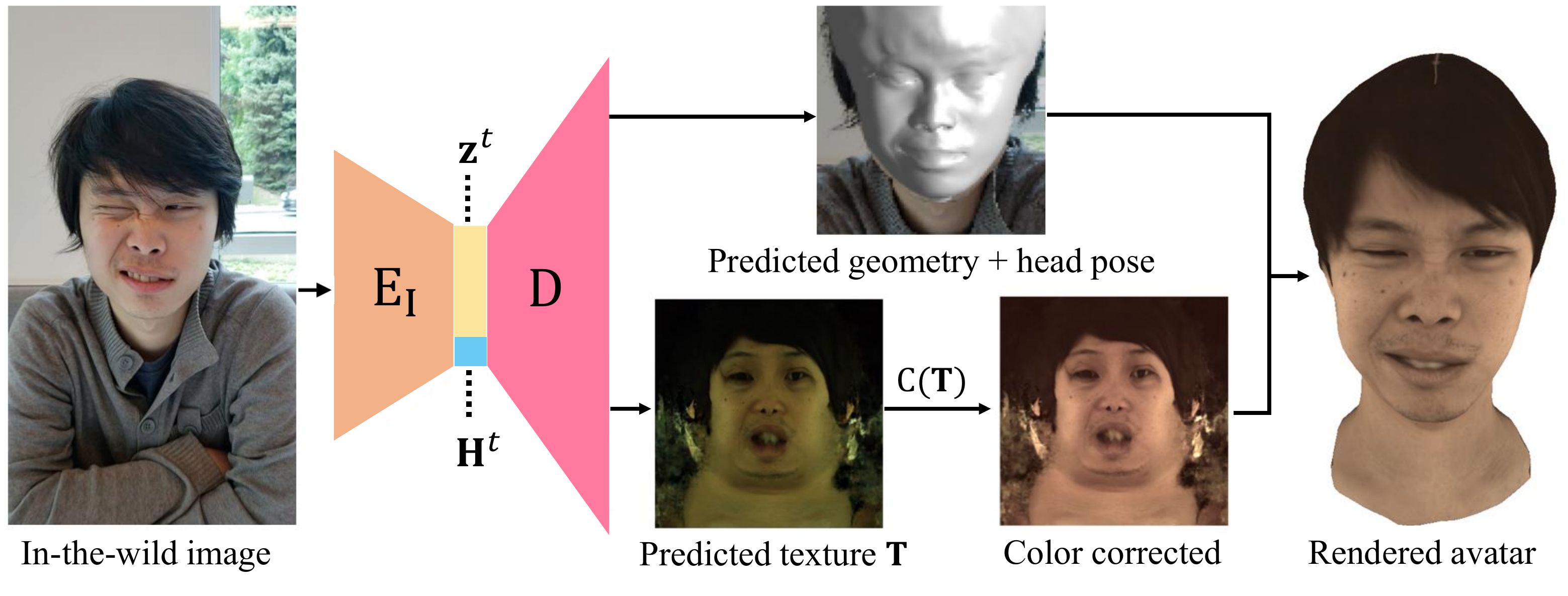}
	\end{center}	
	\vspace{-1mm}
	\caption{Proposed method during testing phase. 
	}
	\label{fig:prediction}
\end{figure}
\vspace{-1mm}
\subsection{Testing Phase}
\vspace{-1mm}
Figure~\ref{fig:prediction} depicts the steps required during the testing phase of our network,
which is simply a feed-forward pass through the adapted $\CodecEncoder_\Image$ and
the estimated color correction function $C$. Note that 
$\widetilde{\mathbf{T}}$ 
and 
the landmark detection 
are no longer required.
Therefore, the timing of the network is still exactly the same as the original network
except for the additional color correction, which itself is simple and  fast.

\section{Experiments}
\label{sec:Experiments}
To demonstrate the effectiveness of our proposed self-supervised domain adaptation method
for high-fidelity 3D face tracking, we perform both quantitative and qualitative analysis.
Though qualitative analysis is relatively straight forward, quantitative analysis
for evaluating the accuracy and stability of tracking results requires a high-resolution 
in-the-wild video dataset with ground-truth 3D meshes,
which unfortunately is difficult to collect because scanning high quality 3D facial scans
usually requires being in a special lab environment with controlled settings.
Thus quantitative analysis of recent 3D face tracking methods such as 
\cite{tewari2017self, tewari2017mofa} are limited to static image datasets \cite{cao2014facewarehouse},
or video sequences shot in a controlled environment \cite{valgaerts2012lightweight}.
Therefore, in light of the aforementioned limitations, 
we collected a new dataset and devised two metrics
for quantitatively evaluating 3D face tracking performance.


\noindent
\textbf{Evaluation metrics}:
We employ two metrics, accuracy and temporal stability, which are denoted as "Reprojection" and "Temporal" in Table~\ref{tb:compare}, respectively.
For accuracy, since we do not have ground truth 3D meshes for in-the-wild data,
we utilize average 2D landmark reprojection error as a proxy for the accuracy of the predicted 3D geometry. 
First, a 3D point corresponding to a 2D landmark is projected into 2D, 
and then the Euclidean distance between the reprojected point and ground truth 2D point is computed.
For temporal stability, we propose a smoothness metric as 
\begin{align}
\frac{1}{G}\sum_{i=1}^{G}\frac{\left\|\mathbf{G}^{t+1}_i-\mathbf{G}^{t}_i\right\|_2+\left\|\mathbf{G}^t_i-\mathbf{G}^{t-1}_i\right\|_2}{\left\|\mathbf{G}^{t+1}_{i}-\mathbf{G}^{t-1}_{i}\right\|_2},
\label{cr:temporal}
\end{align}
where $\mathbf{G}^t_i$ corresponds to the 3D location of vertex $i$ at time $t$.
This metric assumes that the vertices of the 3D mesh should move on a straight line over the 
course of three frames, thus unstable or jittering predictions will lead to higher (worse) score. The lowest (best) metric score is 1.

\noindent
\textbf{Dataset collection and annotation}:
We recorded 1920$\times$1080 resolution facial performance data in the wild for four different identities.
Recording environments include indoor, outdoor, plain background and cluttered background
under various lighting conditions.

150 frames of facial performance data were annotated for each of the 4 identities.
For each frame, we annotate on the person's face 5 salient landmarks that do \textbf{not}
correspond to any typical facial landmark such as eye corners and mouth corners that can be detected by our landmark detector. 
These points are selected because our domain adaptation method already optimizes for facial landmark reprojection consistency, 
so our evaluation metric should use a separate set of landmarks for evaluation.
Therefore, we focus on annotating salient personalized landmarks,
such as pimples or moles on a person's face, which can be
easily identified and accurately annotated by a human. In this way, our annotations enable us to measure performance of tracking in regions
where there are no generic facial landmarks and provide a more accurate measure of tracking performance.

\textbf{Implementation Details}:
DAMs~\cite{Lombardi_SIGGRAPH2018} are first created for all four identities from multi-view images captured in a lighting-controlled environment, and our \RegressorName\ is newly trained for each identity. Our proposed self-supervised domain adaptation method
is then applied to videos of the four identities in a different lighting and background environment. For DAM, the unwrapped texture resolution is $T=1024$, and the geometry had $G=7306$ vertices.
We train the \RegressorName$\,$ with Stochastic Gradient Decent (SGD). 
The face is cropped and resized to 256$\times$256 image and given to $\CodecEncoder_\Image$.
During the self-supervised domain adaptation, the related parameters are set 
to $\lambda_{\CFTC}=100,\ \lambda_{\MOTC}=100,\ \lambda_{\FLRC}=1$. 

\begin{table}[t]
	\centering
	\caption{Evaluation on in-the-wild dataset. ``Ours w/o DA'' represents $\CodecEncoder_\Image$ before doing any domain adaptation.} 
	\vspace{-3mm}
 	\begin{tabular}[t]{p{0.8cm}|p{0.8cm}|p{7.2mm}p{7.2mm}p{7.2mm}p{7.2mm}||p{7.2mm}}   
		\toprule[1.5pt]  
		\\[-3ex]
		&{}
        &\scriptsize{Subject1}
        &\scriptsize{Subject2}
        &\scriptsize{Subject3}
        &\scriptsize{Subject4}
        &\scriptsize{Average}

     	\\[-3ex]\midrule[1.5pt]  \\[-3.8ex]
     	
		
         \parbox{0.5cm}{\vspace{5mm}\centering\scriptsize{HPEN}}
		&\parbox{0.4cm}{\centering} 
		{\hspace{-1.8mm}\tiny{Temporal}} 
		&\scriptsize{1.5197}&\scriptsize{1.2951}&\scriptsize{{{1.8206}}}&\scriptsize{{{1.3559}}}&{\scriptsize{1.4978}}
		\\[-1ex]


		\parbox{0.5cm}{\centering\scriptsize{}} 
			&\parbox{0.4cm}{\centering} 
		{\hspace{-1.8mm}\tiny{Reprojection}} 
		&\scriptsize{8.8075}&\scriptsize{\textbf{5.5475}}&\scriptsize{{{13.3823}}}&\scriptsize{{{10.4688}}}&{\scriptsize{9.5515}}
	    \\[-2ex]  \cmidrule{1-7}\\[-4ex]
	
	
		\parbox{0.5cm}{\vspace{5mm}\centering\scriptsize{3DDFA}} &\parbox{0.4cm}{\centering} 
		{\hspace{-1.8mm}\tiny{Temporal}} 
		&\scriptsize{1.5503}&\scriptsize{1.4500}&\scriptsize{1.8608}&\scriptsize{1.5139}&{\scriptsize{1.5938}}\\[-1ex]
		
		&\parbox{0.4cm}{\centering} 
		{\hspace{-1.8mm}\tiny{Reprojection}} 
		&\scriptsize{14.1171}&\scriptsize{10.2568}&\scriptsize{{{21.5077}}}&\scriptsize{{{18.1647}}}&{\scriptsize{16.011}}
		\\[-2ex]
        \\[-2.4ex]  \cmidrule{1-7}\\[-4ex]
		

		\parbox{0.5cm}{\vspace{5mm}\scriptsize{PRNet}} 
			&\parbox{0.4cm}{\centering} 
		{\hspace{-1.6mm}\tiny{Temporal}} 
		&\scriptsize{1.5551}&\scriptsize{1.3701}&\scriptsize{{1.5700}}&\scriptsize{1.4973}&{\scriptsize{1.4981}}
		 \\[-1ex]

		\parbox{0.5cm}{\centering\scriptsize{}} &\parbox{0.4cm}{\centering} 
		{\hspace{-1.8mm}\tiny{Reprojection}} 
		&\scriptsize{8.4867}&\scriptsize{7.2522}&\scriptsize{14.052}&\scriptsize{{{9.6586}}}&{\scriptsize{9.8624}}
        \\[-2ex]
        \\[-1ex]  \cmidrule{1-7}\\[-3.4ex]
		
        \parbox{0.5cm}{\vspace{1mm}\scriptsize{Ours}} 
		&\parbox{0.4cm}{\centering} 
		{\hspace{-1.8mm}\tiny{Temporal}} 
		&\scriptsize{1.4106}&\scriptsize{1.2476}&\scriptsize{1.8322}&\scriptsize{1.4169}&{\scriptsize{1.4768}}
		\\[-0.2ex]


    	\parbox{0.5cm}{\vspace{0mm}\hspace{0mm}\scriptsize{w/o~DA}} 
			&\parbox{0.4cm}{\centering} 
		{\hspace{-1.8mm}\tiny{Reprojection}} 
		&{\scriptsize{6.2171}}&{\scriptsize{7.4914}}&{\scriptsize{10.9225}}&{\scriptsize{{{9.5953}}}}&{{\scriptsize{8.5566}}}
		\\[-2ex]
        \\[-3.5ex]  \cmidrule{1-7}\\[-3.5ex]
		
		
	        \parbox{0.5cm}{\vspace{2mm}\scriptsize{Ours}} 
		&\parbox{0.4cm}{\centering} 
		{\hspace{-1.8mm}\tiny{Temporal}} 
		&\scriptsize{1.3624}&\scriptsize{1.3274}&\scriptsize{1.6583}&\scriptsize{1.132}&{\scriptsize{1.3700}}
		\\[-0.2ex]


    	\parbox{0.5cm}{\vspace{0mm}\hspace{-3mm}\scriptsize{w/~$L_{\textrm{FLRC}}$}} 
			&\parbox{0.4cm}{\centering} 
		{\hspace{-1.8mm}\tiny{Reprojection}} 
		&{\scriptsize{5.7558}}&{\scriptsize{6.982}}&{\scriptsize{10.1258}}&{\scriptsize{{{7.5230}}}}&{{\scriptsize{7.5960}}}
		\\[-2ex]
        \\[-3.5ex]  \cmidrule{1-7}\\[-4ex]
		
		
		%
		\parbox{0.5cm}{\vspace{5mm}\centering\scriptsize{Ours}} 
			&\parbox{0.4cm}{\centering} 
		{\hspace{-1.6mm}\tiny{Temporal}} 
		&\scriptsize{\textbf{1.1299}}&\textbf{\scriptsize{1.0498}}&{\scriptsize{\textbf{1.2934}}}&\scriptsize{\textbf{1.0915}}&{\textbf{\scriptsize{1.1412}}}
        \\[-1.2ex]

		
        &\parbox{0.4cm}{\centering} 
		{\hspace{-1.8mm}\tiny{Reprojection}} 
		&\textbf{\scriptsize{5.5689}}&{\scriptsize{6.7281}}&\textbf{\scriptsize{9.6015}}&\textbf{\scriptsize{{{7.1368}}}}&{\textbf{\scriptsize{7.2588}}}
		\\[-0.5ex] 
		\bottomrule[1.5pt]
		
	\end{tabular}     

	\label{tb:compare}
\end{table}

\begin{figure}[t]
\vspace{-3mm}
\includegraphics[width=0.98\linewidth]{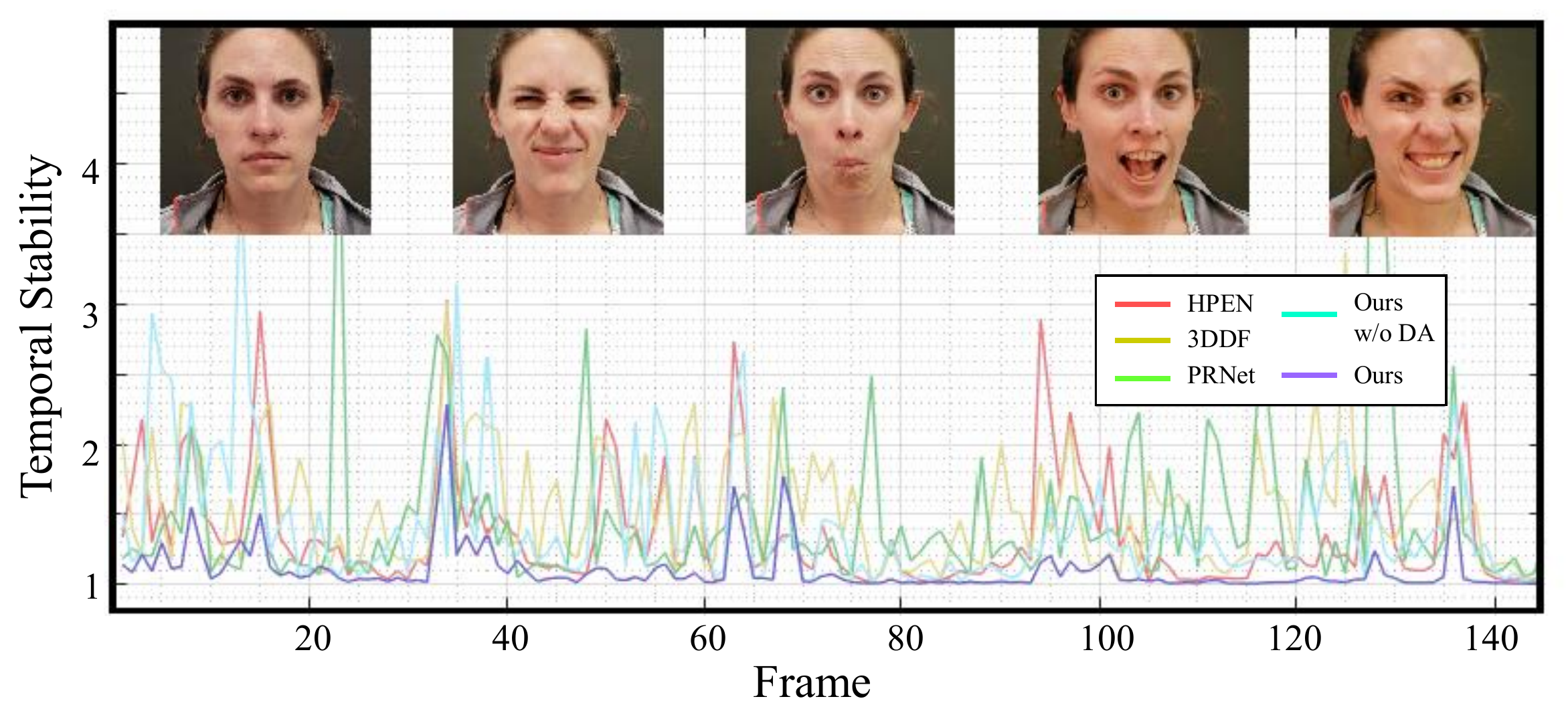}
	\vspace{-4mm}
	\caption{Temporal stability graph for subject 4. Note that smaller stability score means
	more stable results.}
	\label{fig:self_eval3}
\end{figure}
\subsection{Results on In-the-wild Dataset}
\label{sec:exp_inthewild}
We compare our method against three state-of-the-art baselines:
\textbf{HPEN}~\cite{zhu2015high}: 3DMM fitting based on landmarks, 
\textbf{3DDFA}~\cite{zhu2016face}: 3DMM fitting based on landmarks and dense correspondence, 
and \textbf{PRNet}~\cite{feng2018joint}: 3DMM fitting based on the direct depth regression map. 
The system input image size is 256$\times$256 except for \textbf{3DDFA} (100$\times$100). 
We also add our method without domain adaptation (\textbf{Ours w/o DA}) and only with facial landmark reprojection consistency (\textbf{Ours w/ $L_{\textrm{FLRC}}$}).
As shown in Table~\ref{tb:compare}, the proposed domain adaptation consistently increases the performance
of the our model without domain adaptation for all 4 subjects.
In terms of stability, the proposed domain adaptation method improves our model  by 22\% relative.
Particularly, we are able to achieve 1.05 stability score for subject 2, which is close to the lowest possible stability score (1.0).
This demonstrates the effectiveness of our proposed method. 
For the other baselines, our model without the domain adaptation already outperforms them in terms of geometry. 
This may be because our model is pre-trained with many pairs of $(\Image, \textbf{H}, \code)$ training data, while the baselines were used out of the box.
But on the other hand, all baselines including \textbf{Ours w/o DA} perform similarly in terms of stability (between 1.45-1.60), but
our domain adaptation method is able to improve it to 1.14, which clearly demonstrates the effectiveness of 
our method.


Figure~\ref{fig:self_eval3} visualizes the temporal stability metric for all the different methods
for a single sequence. Our method has a consistently better (\ie, smaller)  stability score 
than all the other methods for nearly all the frames, and demonstrates not only the
effectiveness, but also the reliability and robustness of our method for in-the-wild sequences.

\begin{figure}[t]
\vspace{-4mm}
	\centering
	\includegraphics[width=0.98\linewidth]{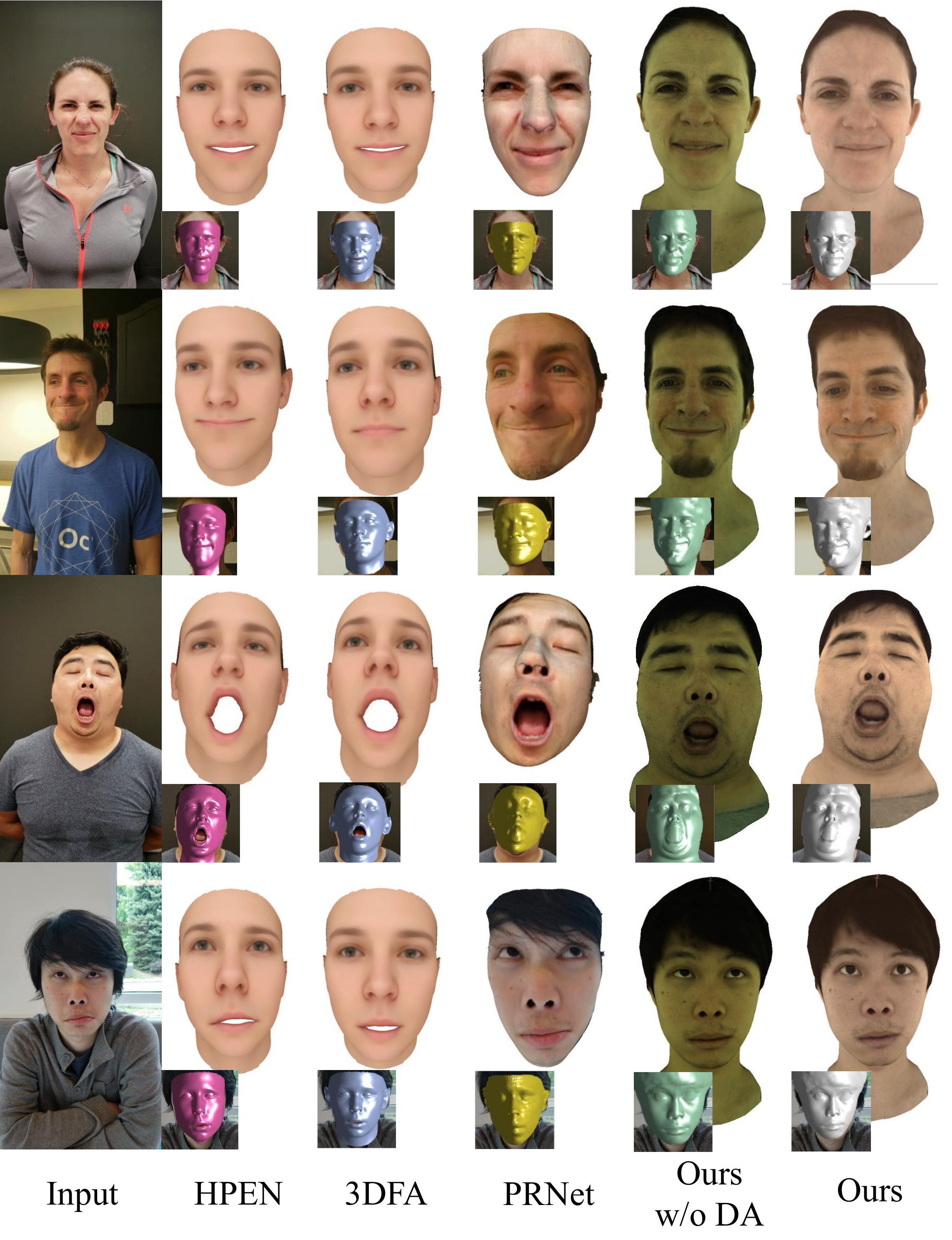}
	\caption{Qualitative comparisons with baseline methods. }
	\label{fig:qualitative2}
	\vspace{-4mm}
\end{figure}
Figure~\ref{fig:qualitative2} shows qualitative comparisons with baselines. 
Overall, our face tracking results most closely
resemble the input facial configuration, especially for the eyes and the mouth. For example, in the second row, the baselines erroneously predicted that the person's mouth is opened, while our method correctly predicted that the person's mouth is closed. We can also clearly see the effectiveness of our color correction approach, which is able to correct the relatively green-looking face to better match to the appearance in the input.

\begin{figure}[t]
\vspace{-3mm}
\centering
\includegraphics[width=0.98\linewidth]{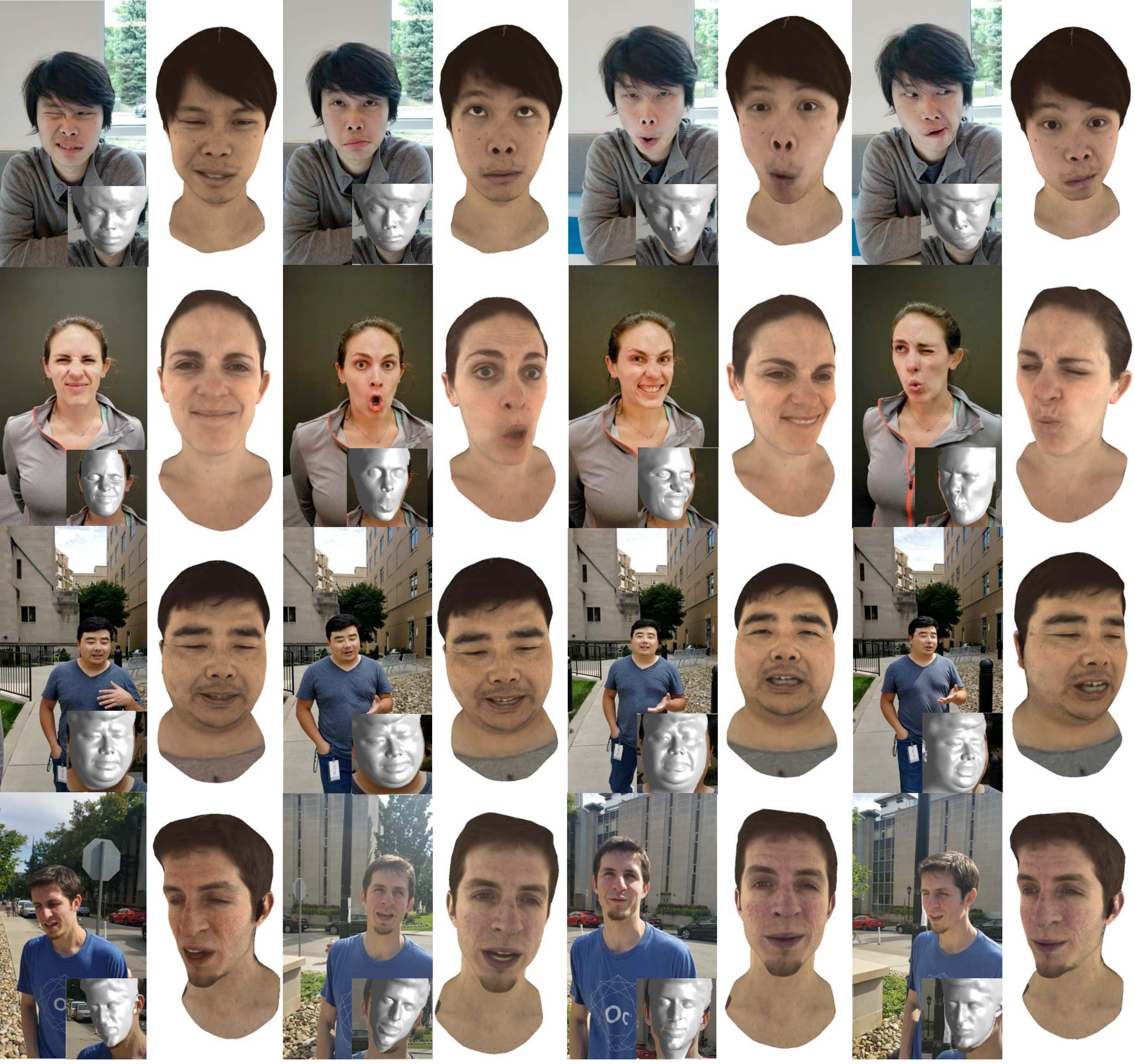}
	\vspace{-2mm}
	\caption{Visualization of 3D face tracking for in-the-wild video. For each input image, we show in the bottom right corner the predicted geometry overlaid on top of the face,
	and the predicted color corrected face.
}
	\label{fig:4ppl_qualitative}
\end{figure}

Figure~\ref{fig:4ppl_qualitative} shows the visualization of our in-the-wild face tracking results.
Our method is able to track complex motion in many different backgrounds, head pose,
and lighting conditions that are difficult to approximate with spherical harmonics such as hard shadow. Our method is also able to adapt to the white-balance
of the current scene. Note that the gaze direction is also tracked for most cases.

\subsection{Ablation Studies} 
To gain more insight to our model, we performed the following ablation experiments. 
\vspace{-4mm}
\subsubsection{Evaluation of {\textbf{\RegressorName}} Structure}\label{ablation_network}
\vspace{-2mm}
To validate the performance gain of each component on our regression network, we compare \RegressorName\ against  three baseline networks: \textbf{VGG+Skip+Key} denotes \RegressorName, which uses VGGNet, multi-level features (skip connections), and landmarks from HourglassNet. \textbf{VGG+Skip}: landmarks guidance is removed. \textbf{VGG}: Multi-level features (skip connection) are further removed and only deep features are used for regression. \textbf{VGG Scratch} has the same structure with \textbf{VGG} but it is trained from scratch. For other settings which use \textbf{VGG}, pre-trained VGG-16 features are used, and the VGG portion of the network is not updated during training. The models are tested on unseen test datasets where the vertex-wise dense ground-truth is available. Three metrics are employed to evaluate performance:  (1) accuracy for geometry is computed by Euclidean distance between predicted and ground-truth 3D vertices, (2) accuracy for texture is calculated by pixel intensity difference between predicted and ground-truth texture, and (3) the temporal stability is measured in the same way as Eq.~\ref{cr:temporal}.

\begin{figure}[t]
\vspace{-3mm}
		\begin{center}
\hspace{-7mm}
\includegraphics[width=3.5in]{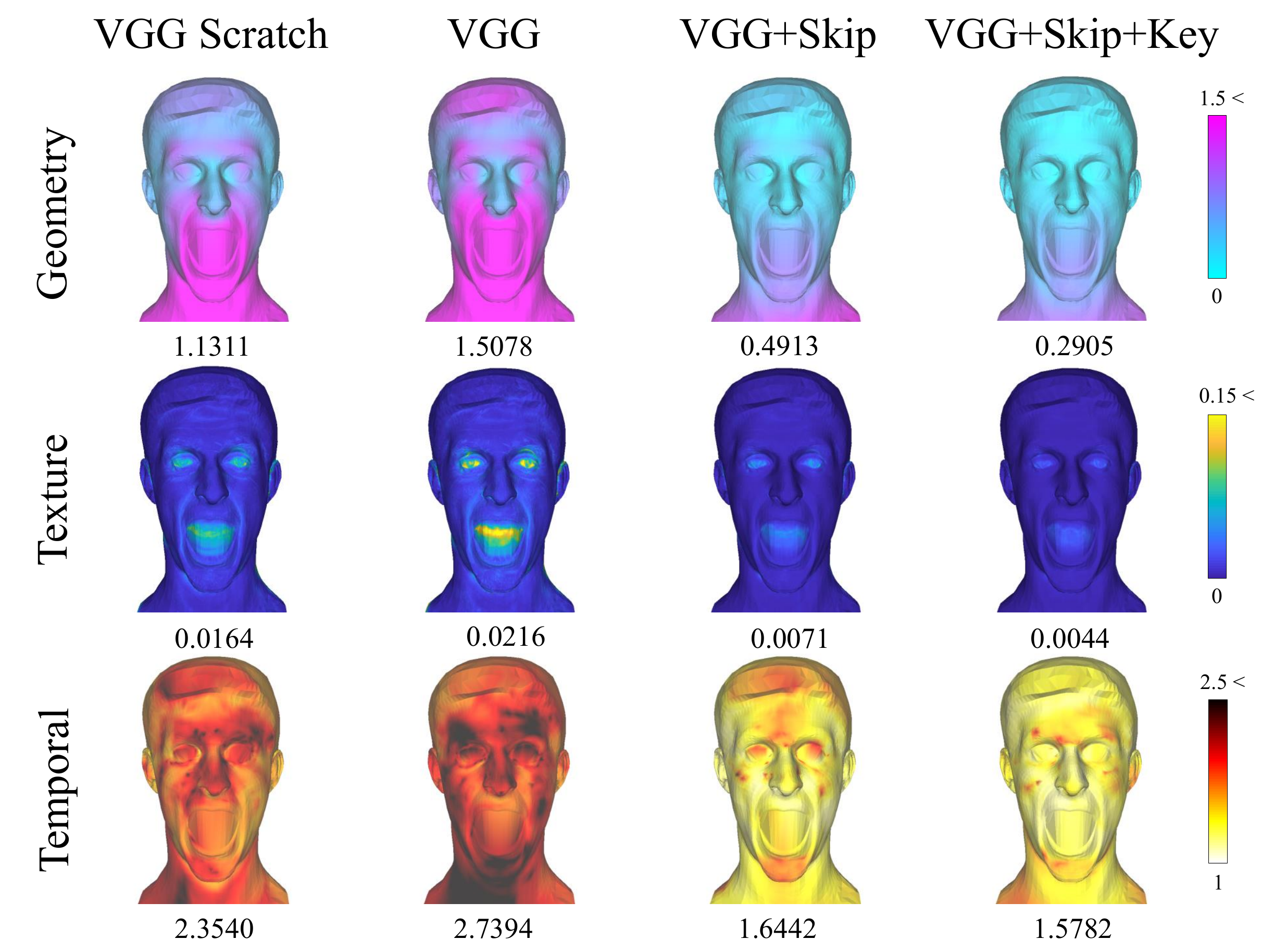}
\end{center}
 	\vspace{-3mm}
	\caption{Ablation test on \RegressorName\ with a representative subject. The vertex-wise error is visualized with the associated average score for subject 1.}
	\label{fig:ablation_dense}
\end{figure}

The average scores with respect to the four test subjects are reported in Table~\ref{tb:ablation_network}, and the representative subject results are visualized in Figure~\ref{fig:ablation_dense}. We observe that multi-level features (\textbf{VGG+Skip}) significantly improves performance over \textbf{VGG}, while adding keypoints (\textbf{VGG+Skip+Key}) further improves performance. 
\textbf{VGG} seems to lack of capacity to directly regress the latent parameters with only pre-trained deep features which are not updated. More ablation studies (\eg, tests on view consistency and robustness to the synthetic visual perturbation) on \RegressorName\ are described in the supplementary manuscript.



\begin{table}[t]
	\caption{Ablation test on \RegressorName. The average score with respect to all subjects are reported.}\vspace{-3mm} 

	\centering
		
 	\begin{tabular}[t]{p{0.9cm}|p{11.9mm}p{11.9mm}p{14mm}p{14mm}}   
		\toprule[1.5pt]  
		\\[-3ex]
		&\scriptsize{VGG~Scratch}
		&\hspace{2mm}\scriptsize{VGG}
		&\hspace{0mm}\scriptsize{VGG+Skip}
		&\hspace{-3mm}\scriptsize{VGG+Skip+Key}

     	\\[-3ex]\midrule[1.5pt]  \\[-3ex]
		\parbox{0.5cm}{\centering\scriptsize{Geometry}} 
		&{\scriptsize{\hspace{2.5mm}1.011}}&{\scriptsize{\hspace{2.5mm}1.481}}&\scriptsize{\hspace{2.5mm}0.411}&\scriptsize{\hspace{1.5mm}\textbf{0.315}}
		\\[-0ex] \cmidrule{1-5}\\[-3.2ex]
		\parbox{0.5cm}{\centering\scriptsize{Texture}} 
		&{\scriptsize{\hspace{2.5mm}0.016}}&{\scriptsize{\hspace{2.5mm}0.027}}&\scriptsize{\hspace{2.5mm}0.007}&\scriptsize{\hspace{1.5mm}\textbf{0.004}}
		\\[-0ex] \cmidrule{1-5}\\[-3.2ex]
     	\parbox{0.5cm}{\centering\scriptsize{Temporal}} 
		&{\scriptsize{\hspace{2.5mm}2.143}}&{\scriptsize{\hspace{2.5mm}3.138}}&\scriptsize{\hspace{2.5mm}1.499}&\scriptsize{\hspace{1.5mm}\textbf{1.446}}

		\\[-0.1ex]  
		\bottomrule[1.5pt]
	\end{tabular}     
	\label{tb:ablation_network}
\end{table}

\vspace{-2mm}
\subsubsection{Effect of Image Resolution}
\begin{figure}[t]
\hspace{-7mm}
		\begin{center}
\includegraphics[width=3.2in]{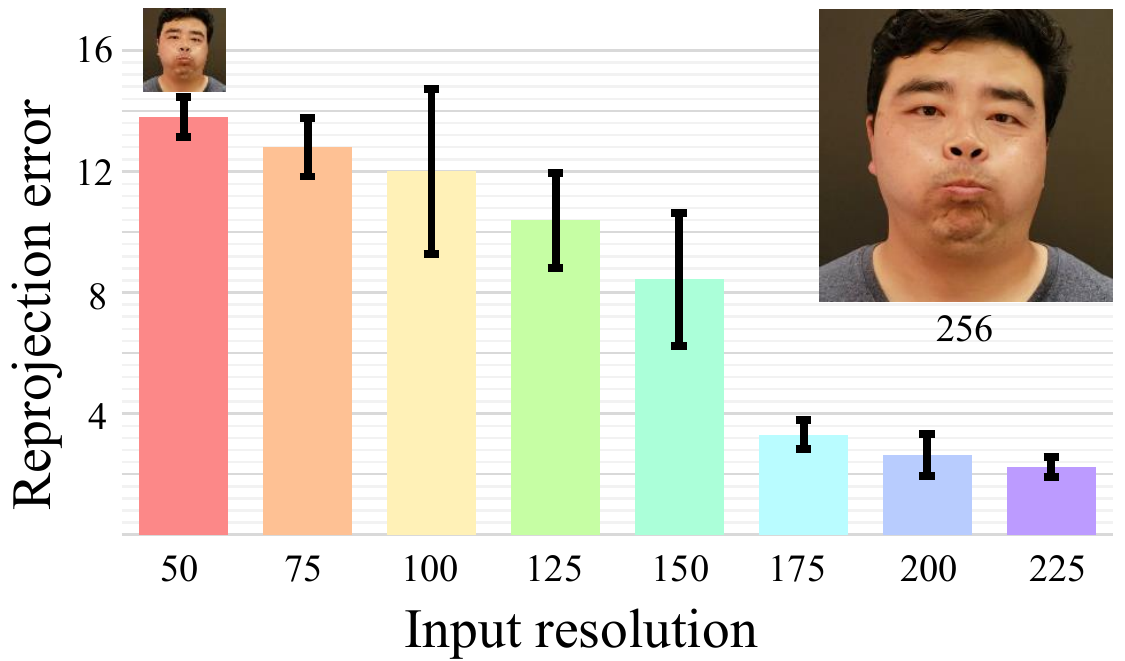}
\end{center}
 	\vspace{-3mm}
	\caption{Ablation studies on the performance degradation under various input resolution.}
	\label{fig:ablation2}
\end{figure}

\vspace{-2mm}
The cropped image resolution plays a key role in the accuracy of face tracking. In this experiment, we quantify the performance degradation according to the resolution using relative reprojection error metric. 
Relative reprojection error is computed by comparing the 2D reprojected vertices location of the estimated geometry from different resolution images
with the one of the gold-standard geometry, which is the geometry acquired when using the highest
image resolution 256$\times$256.
Figure~\ref{fig:ablation2} shows the results. Until 175$\times$175, we achieve average error 
less than 4~pixel-error, but performance degrades significantly as the resolution becomes further smaller.

\vspace{-2mm}
\subsection{Limitations}
\vspace{-2mm}
There are two main limitations to the proposed approach.
The first limitation is that our method assumes that a person-specific DAM already exists for the person to be tracked, as our method takes the DAM as input.
The second limitation is that our MOTC color correction cannot handle complex lighting and specularities.
For example, in Figure~\ref{fig:4ppl_qualitative} first row first image, a portion 
of the face is brighter due to the sun, but since we only have a global color correction matrix
for color correction, the sun's effect could not be captured and thus not reflected in the output.
\vspace{-1mm}
\section{Conclusion}
\label{sec:Conclusion}
\vspace{-1mm}

We proposed a deep neural network that predicts the intermediate representation and head pose of a high-fidelity 3D face model from a single image and its self-supervised domain adaptation method, thus enabling high-quality facial performance tracking from a monocular video in-the-wild. Our domain adaptation method leverages the assumption that the textures of a face over two consecutive frames should
not change drastically, and this assumption enables us to extract supervision from unlabeled 
in-the-wild video frames to fine-tune the existing face tracker. The results demonstrated that the proposed method not only improves face-tracking accuracy, but also
the stability of tracking.








\section*{Acknowledgement}
This work was partially supported by NSF Grant IIS 1755895.

\bibliographystyle{ieee_fullname}
\bibliography{ref}


\twocolumn[{%
  \newpage
  \null
  \vskip .375in
  \begin{center}
      {\Large \bf Supplementary Material:\\
Self-Supervised Adaptation of High-Fidelity Face Models for \\ Monocular Performance Tracking \par}
      \vspace*{24pt}
      {
      \large
      \lineskip .5em
      \begin{tabular}[t]{c}
         \ifcvprfinal Jae Shin Yoon$^\dagger$
\hspace{10mm}Takaaki Shiratori$^\ddagger$
\hspace{10mm}Shoou-I Yu$^\ddagger$
\hspace{10mm}Hyun Soo Park$^\dagger$
\\
\hspace{0mm}$^\dagger$University of Minnesota
\hspace{20mm}$^\ddagger$Facebook Reality Labs
\\
{\tt\small \hspace{8mm}\{jsyoon, hspark\}@umn.edu \hspace{6mm} \{tshiratori, shoou-i.yu\}@fb.com}
\\

         \vspace*{1pt}\\
      \end{tabular}
      \par
      }
      \vskip .5em
      \vspace*{12pt}
  \end{center}
  
  \vspace{-5mm}

  	\begin{center}
\includegraphics[width=7.2in]{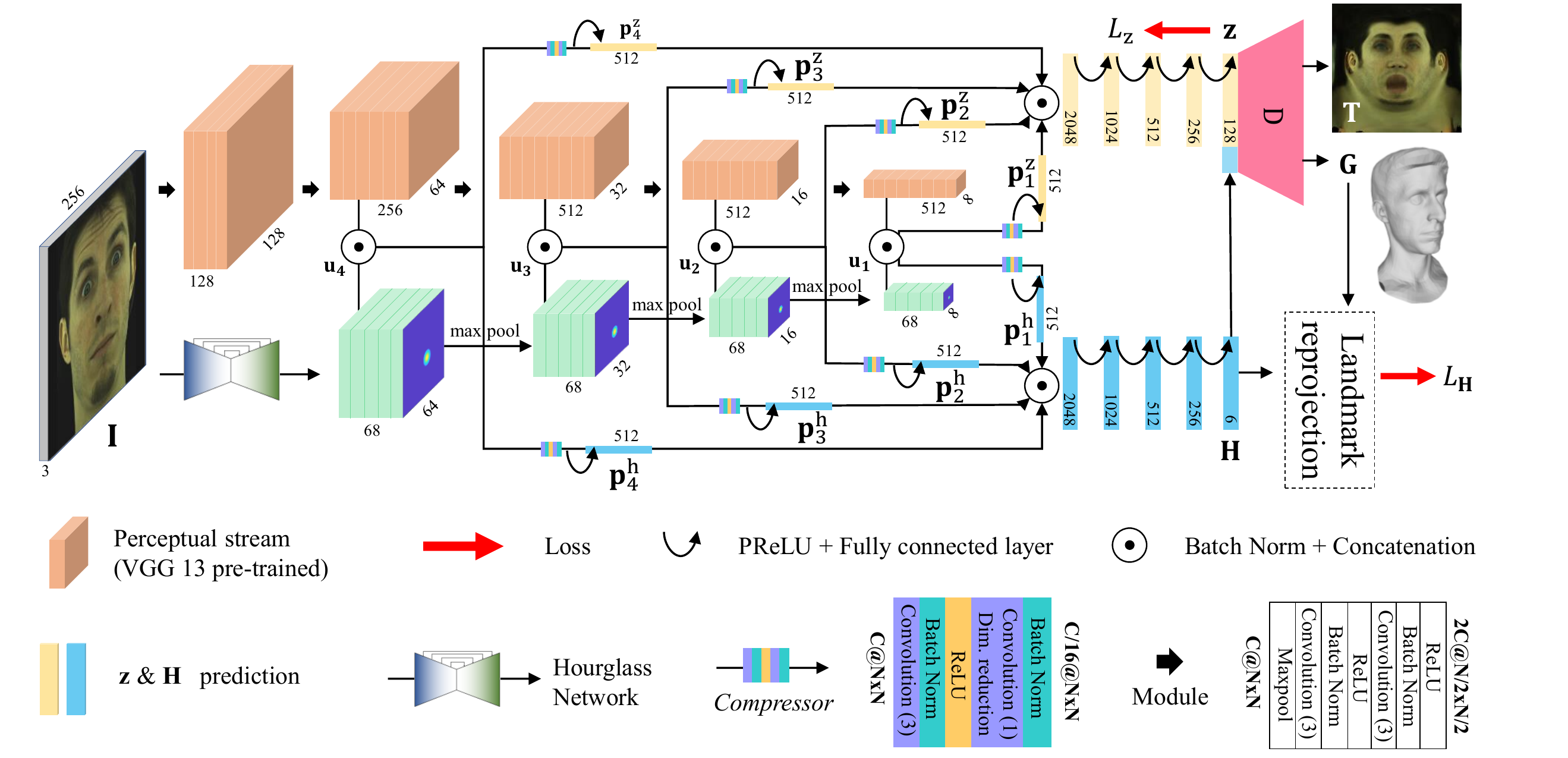}
	\captionof{figure}{\RegressorName\ directly regresses the latent facial state codes $\mathbf{z}$ and headpose $\mathbf{H}$ from a face image $\textbf{I}$, and the pre-trained decoder $\CodecDecoder$ generates full 3D face geometry \textbf{G} and high resolution texture \textbf{T}.}
	\label{fig:network}
\end{center}
  \vspace{5mm}
}]
\setcounter{section}{0}
\def\thesection{\Alph{section}}
\renewcommand{\thesubsection}{\thesection.\arabic{subsection}}

In the supplementary materials, we provide details on the architecture of \RegressorName\ in Section~\ref{sec:Method}, and the additional ablation studies on \RegressorName\ will be followed in Section~\ref{exp:supple}. 


\section{\RegressorName}
\label{sec:Method}
\vspace{-3mm}
In this section, we detail the architecture of \RegressorName. 

Other than only utilizing self-supervised domain adaptation to overcome domain differences, we also explored different networks which could lead to the most domain invariance.
Namely, we utilized a combination of a pre-trained VGGNet~\cite{simonyan2014very} and HourglassNet~\cite{newell2016stacked} to achieve better domain invariance. More details are in Section~\ref{sec:unified}, and the property of domain invariant features are validated in Section~\ref{exp:supple}.

Domain specific layers are still necessary to complete our tasks, but thanks to the domain invariant features already extracted by VGGNet and HourglassNet, the domain specific layers can have less parameters thus they are easier to train. We use a combination of deep and shallow features to achieve better performance. More details are in Section~\ref{sec:regress}. 




\subsection{Inputs and Outputs}
Given a cropped input face image $\mathbf{I}\in \mathbb{R}^{256\times256\times3}$, the {\RegressorName}\ directly predicts the low-dimensional facial state codes $\mathbf{z}\in \mathbb{R}^{128}$, and a set of head pose parameters $\mathbf{H}\in \mathbb{R}^6=\{f,\ r_x,\ r_y,\ r_z,\ t_x,\ t_y \}$, where $\mathbf{f}=\{f\},\ \mathbf{r}=\{r_x,\ r_y,\ r_z\},\ \mathbf{t}=\{t_x,\ t_y\}$ are focal length scale, Euler angle, and 2D translation respectively. The pre-trained decoder $\CodecDecoder$ decodes $[\mathbf{z}^{\mathsf{T}}, \mathbf{H}^{\mathsf{T}}]$ to generate high fidelity 3D face geometry $\mathbf{G}\in \mathbb{R}^{7306\times3}$ and view dependent texture map $\mathbf{T}\in \mathbb{R}^{1024\times1024\times3}$. Note that, we are using the same decoder with~\cite{Lombardi_SIGGRAPH2018}, while we replace its encoder network $\text{E}_X$ with our {\RegressorName}. 

\subsection{Domain Invariant Multi-level Unified Features}\label{sec:unified} 
Given an input image $\mathbf{I}$, {\RegressorName} extracts the features from two-stream networks: VGGNet and HourglassNet. VGGNet captures perceptual information such as facial details or shape, while HourglassNet guides "where to look" by providing facial geometry features, e.g. facial landmark heatmaps. We complete the multi-level unified features $\mathbf{u}_{l}\in \mathbb{R}^{(32*2^{l})\times(32*2^{l})\times ch_{l}}$ by concatenating the two-stream features, where $l=\{4,3,2,1\}$ denotes the feature depth-level and the associated channel size is $ch_{l}\in~CH=\{324, 580, 580, 580\}$. Here, we simply max-pool the output from HourglassNet to make the feature size equal to each level of VGG feature. The feature scale inconsistency between two different networks (VGGNet and HourglassNet) is resolved by normalization layer before concatenation. Our multi-level unified features are more domain (color, illumination, or head pose) invariant by learning from domain generalized datasets~\cite{imagenet_cvpr09, sagonas2016300}. Note that, the pre-learned weights on the two-stream networks are fixed in the following training steps such that we prevent {\RegressorName} from being domain specific. 


\subsection{Latent Parameter Regression} \label{sec:regress} 
\ Inspired by many recent papers~\cite{tao2016siamese, yoon2017pixel} which have proposed the use of combination of deep and shallow features to capture semantic-level information and local appearance details at the same time, we concatenate feature vectors from each depth level $\mathbf{p}_{4..1}^{z},\ \mathbf{p}_{4..1}^{h}\in \mathbb{R}^{512}$, which are encoded from $\mathbf{u}_{4..1}$, and they are respectively regressed to $\mathbf{z}$ and $\mathbf{H}$ using several fully connected layers. Here, however, it requires very heavy computational costs for converting three-dimensional features $\mathbf{u}_{l}$ to single dimensional one $\mathbf{p}_{l}^{z,h}$ in a fully connected way. Similar to~\cite{yoon2017pixel}, we alleviate this bottleneck by channel-wise feature compression of $\mathbf{u}_{l}$ to one-sixteenth of its original channel size using two convolutional layers as described as \textit{Compressor} layer in Figure~\ref{fig:network}.

\begin{table}[t]
	\caption{Ablation studies on \RegressorName.}\vspace{-3mm} 

	\centering
		
 	\begin{tabular}[t]{p{1.2cm}|p{0.6cm}|p{4.2mm}p{4.2mm}p{4.2mm}p{2.2mm}p{7.2mm}}   
		\toprule[1.5pt]  
		\\[-3ex]
		&{\scriptsize{}}
        &{V}\tiny{iew}
        &{C}\tiny{olor}
        &{L}\tiny{ight}
        &{J}\tiny{itter}
        &{B}\tiny{ackground}
     	\\[-3ex]\midrule[1.5pt]  \\[-3.5ex]
     	

		\parbox{0.5cm}{\vspace{5mm}\centering\scriptsize{}} 
		&\parbox{0.4cm}{\centering} 
    	{\hspace{-1.8mm}\tiny{Geometry}} 
		&\scriptsize{0.607}&\scriptsize{1.485}&\scriptsize{{{1.175}}}&\scriptsize{{{0.983}}}&{\hspace{1mm}\scriptsize{1.285}}
		\\[-2.3ex]


		\parbox{0.5cm}{\vspace{-1mm}\centering\scriptsize{VGG}} 
			&\parbox{0.4cm}{\centering} 
		{\hspace{-1.6mm}\tiny{Headpose}} 
		&\scriptsize{-}&\scriptsize{17.48}&\scriptsize{6.965}&\scriptsize{-}&{\hspace{1mm}\scriptsize{15.84}}
		 \\[-2ex]

	
    \parbox{0.4cm}{\centering\vspace{-1mm}\scriptsize{Scratch}}& 
		{\hspace{-0.7mm}\tiny{Texture}} 
		&\scriptsize{-}&\scriptsize{0.021}&\scriptsize{0.014}&\scriptsize{{{0.016}}}&{\hspace{1mm}\scriptsize{0.015}}
		\\[-1.7ex]  \cmidrule{1-7}\\[-3.2ex]
		
		&\parbox{0.4cm}{\centering} 
		{\hspace{-1.8mm}\tiny{Geometry}} 
		&\scriptsize{1.352}&\scriptsize{1.258}&\scriptsize{{{1.510}}}&\scriptsize{{{1.736}}}&{\hspace{1mm}\scriptsize{1.076}}
		\\[-2ex]


		\parbox{0.5cm}{\vspace{2mm}\scriptsize{VGG}} 
			&\parbox{0.4cm}{\centering} 
		{\hspace{-1.6mm}\tiny{Headpose}} 
		&\scriptsize{-}&\scriptsize{16.61}&\scriptsize{13.98}&\scriptsize{-}&{\hspace{1mm}\scriptsize{16.42}}
		 \\[-2ex]

		\parbox{0.5cm}{\centering\scriptsize{}} &\parbox{0.4cm}{\centering} 
		{\hspace{-0.7mm}\tiny{Texture}} 
		&\scriptsize{-}&\scriptsize{0.020}&\scriptsize{0.021}&\scriptsize{{{0.025}}}&{\hspace{1mm}\scriptsize{0.016}}
		\\[-1.8ex]  \cmidrule{1-7}\\[-3.3ex]
\parbox{0.5cm}{\vspace{4mm}\scriptsize{}} 
		&\parbox{0.4cm}{\centering} 
		{\hspace{-1.8mm}\tiny{Geometry}} 
		&\scriptsize{0.3967}&\scriptsize{0.622}&\scriptsize{{{0.227}}}&\scriptsize{{{1.331}}}&{\hspace{1mm}\scriptsize{0.669}}
		\\[-2ex]


	\parbox{0.5cm}{\vspace{-3mm}\scriptsize{VGG}} 
			&\parbox{0.4cm}{\centering} 
		{\hspace{-1.6mm}\tiny{Headpose}} 
		&\scriptsize{-}&\scriptsize{2.579}&\scriptsize{0.728}&\scriptsize{-}&{\hspace{1mm}\scriptsize{8.750}}
		\\[-2ex]


		\parbox{0.5cm}{\vspace{-2mm}\centering\scriptsize{+Skip}} &\parbox{0.4cm}{\centering} 
		{\hspace{-0.7mm}\tiny{Texture}} 
		&\scriptsize{-}&\scriptsize{0.009}&\scriptsize{0.003}&\scriptsize{{{0.018}}}&{\hspace{1mm}\scriptsize{0.009}}
		\\[-2ex]  \cmidrule{1-7}\\[-3.5ex]
	\parbox{0.5cm}{\vspace{3.5mm}\scriptsize{VGG}} 
		&\parbox{0.4cm}{\centering} 
		{\hspace{-1.8mm}\tiny{Geometry}} 
		&\textbf{\scriptsize{0.255}}&\textbf{\scriptsize{0.505}}&\textbf{\scriptsize{{{0.151}}}}&\textbf{\scriptsize{{{0.896}}}}&{\hspace{1mm}\textbf{\scriptsize{0.417}}}
		\\[-2.7ex]


		\parbox{0.5cm}{\vspace{1.5mm}\centering\scriptsize{+Skip}} 
			&\parbox{0.4cm}{\centering} 
		{\hspace{-1.6mm}\tiny{Headpose}} 
		&\scriptsize{-}&\textbf{\scriptsize{1.676}}&\textbf{\scriptsize{0.684}}&\scriptsize{-}&{\hspace{1mm}\textbf{\scriptsize{8.172}}}
		\\[-2.5ex]

		
		\parbox{0.5cm}{\vspace{-0.5mm}\scriptsize{+Key}} &\parbox{0.4cm}{\centering} 
		{\hspace{-0.7mm}\tiny{Texture}} 
		&{\scriptsize{-}}&\textbf{\scriptsize{0.007}}&\textbf{\scriptsize{0.002}}&\textbf{\scriptsize{{{0.012}}}}&{\hspace{1mm}\textbf{\scriptsize{0.006}}}
		\\[-1.5ex]  
		\bottomrule[1.5pt]
		\bottomrule[1.5pt]
		
	\end{tabular}     
	\label{tb:ablation}
\end{table}

\begin{figure}[t]
\vspace{-4mm}
	\hspace{-3mm}\includegraphics[width=3.5in]{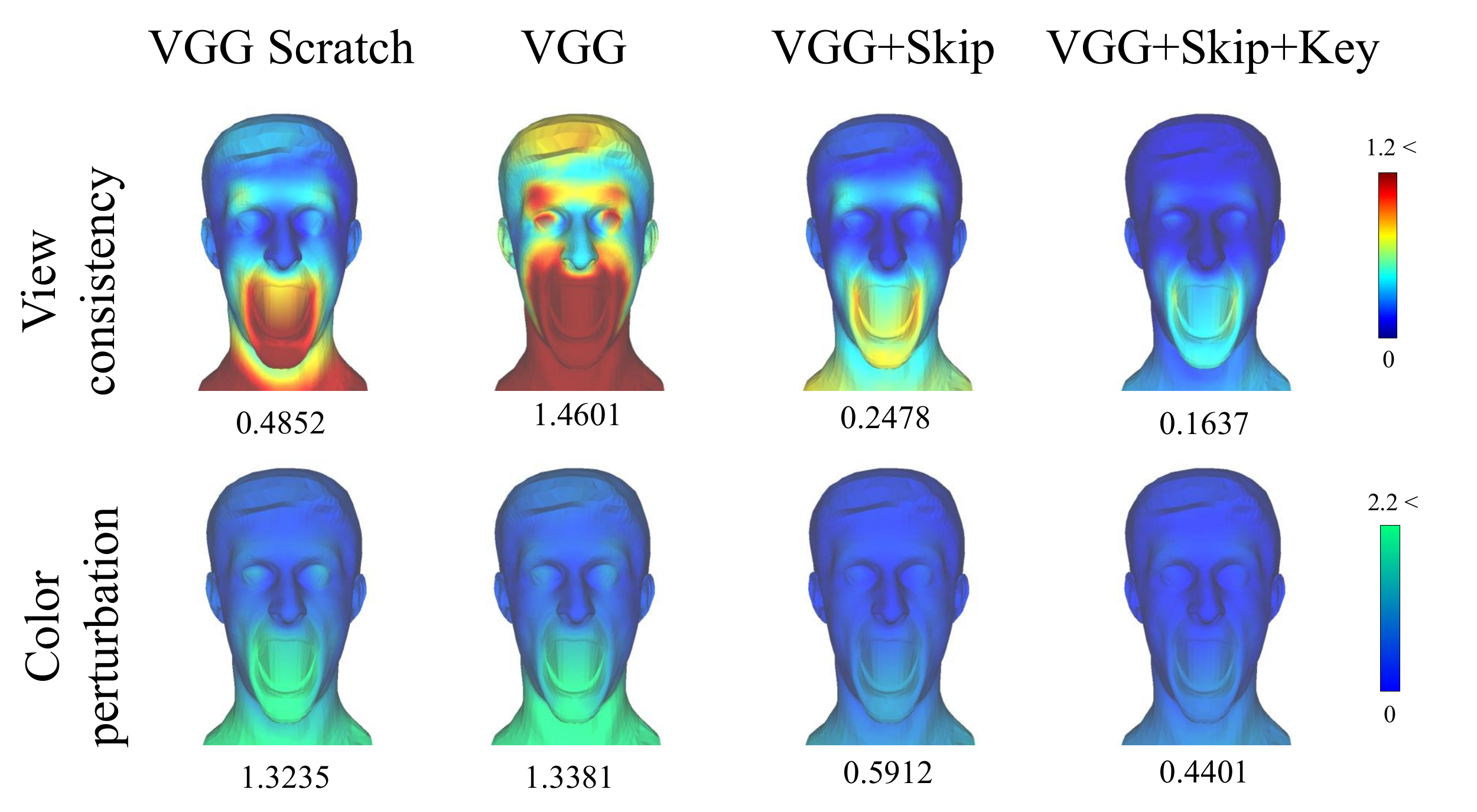}
	\vspace{-8mm}
	\caption{Visualization of the vertex-wise accuracy with a representative subject for ablation studies on view consistency and color sensitivity. The average score is reported for each metric, where the lower score shows the better performance for both scenarios.}
	\label{fig:ablation}
		\vspace{-4mm}
\end{figure}

\section{Ablation Studies on \RegressorName}\label{exp:supple} In Section~\ref{sec:Method}, we introduced the domain and view invariant property of our network. To verify this, we test \RegressorName\ on four different scenarios, \textbf{View}, \textbf{Color}, \textbf{Light}, and \textbf{Jitter}, where the baseline networks are the same with the ones described in Section 4.2.1. 


\textbf{View} represents the test dataset of multiview videos, where they are accurately synchronized and thus \RegressorName\ should predict the same facial local deformation to make the facial configuration consistent across the views. To verify this view consistent prediction ability, we pick the most central camera as a ground-truth view and evaluate the performance of other views. We use simple vertex-wise Euclidean distance between the 3D faces predicted from central view and other views meaning that the lower score shows better consistency. The overall performance is summarized in Table~\ref{tb:ablation} and Figure~\ref{fig:ablation}, where the proposed network outperforms all other baselines. We can further notice that the combination of skip connection and landmark guidance helps the network to figure out the facial geometry configuration when predicting the facial configuration from different views based on the comparison of \textbf{VGG} with \textbf{VGG+Skip} and \textbf{VGG+Skip+Key}. Note that, when evaluating the view consistency, we remove the texture and head pose from a predicted 3D face because they have view dependent property in our system.

\noindent\textbf{Color}, \textbf{Light}, \textbf{Jitter}, and \textbf{Background} represent video sequences which contain synthetic perturbation with random color, gamma, jitters by similarity transformation (scale, rotation, and translation variation), and white dotted background noise. The goal of the test on these sequences is to verify the domain generality. For example, if \RegressorName\ outputs a completely different 3D facial configuration given a perturbed image comparing to the one before the perturbation, then it implies that the network is overfitted to the training data domain. Therefore, we evaluate the performance of \RegressorName\ on the sequence after the perturbation in light of the results from the ones before the perturbation. 
To measure this relative accuracy, we employ three metrics: geometry, texture, and head pose. For geometry and texture, we simply calculate the 3D distance and color difference of the 3D faces. For head pose, we measure the 2D distance between the ground-truth points and the reprojection of the vertices on the 3D face to the input with the predicted head pose. The average scores with respect to the entire test subjects (4 subjects) are reported in Table~\ref{tb:ablation}, and the representative subject results are visualized in Figure~\ref{fig:ablation}. From the comparison of \textbf{VGG Scratch} with \textbf{VGG+Skip+Key}, we can notice that the pre-trained nature of the feature extraction parts (VGGNet and HourglassNet) plays a key role to avoid overfitting from a specific domain. Further, the comparison between \textbf{VGG+Skip} and \textbf{VGG+Skip+Key} implies that the landmark module guides the attention of the network such that it prevents from the network distraction even under the background perturbation.

\end{document}